\documentclass[letterpaper,10pt,conference]{ieeeconf}
\IEEEoverridecommandlockouts
\usepackage[T1]{fontenc}
\usepackage{times}
\usepackage{capt-of}
\usepackage{amsmath,amssymb,booktabs,array,multirow,graphicx,cite}
\usepackage{xcolor,tikz,pgfplots,balance}
\usetikzlibrary{arrows.meta,positioning,calc,fit,patterns}

\definecolor{rblue}{rgb}{0,0.5,1}
\definecolor{awesome}{rgb}{1.0, 0.13, 0.32}
\definecolor{hollywoodcerise}{rgb}{0.96, 0.0, 0.63}
\definecolor{lasallegreen}{rgb}{0.03, 0.47, 0.19}
\definecolor{hanpurple}{rgb}{0.32, 0.09, 0.98}
\definecolor{green(pigment)}{rgb}{0.0, 0.65, 0.31}

\makeatletter
\let\NAT@parse\undefined
\makeatother
\usepackage[pagebackref=false, breaklinks=true, colorlinks, bookmarks=false]{
        hyperref
}
\hypersetup{
        colorlinks=true,
        linkcolor={red},
        citecolor={hanpurple},
        urlcolor={magenta}
}

\pgfplotsset{compat=1.18}
\newcommand{\bench}{LiDAR-Hallu}
\newcommand{\ind}{\mathbf{1}}
\newcommand{\nm}{B4DL}
\newcommand{\mt}{B4DL-Ego}

\title{\LARGE \bf
Do LiDAR Language Models Really Understand\\Spatio-temporal Relationships?}
\author{
Runyi Yang$^{1,*}$ \quad
Murat Akkoyun$^{2,*}$ \quad
Di Wen$^{2}$ \quad
Ruiping Liu$^{2}$ \quad
Yufan Chen$^{2}$ \quad
Junwei Zheng$^{2}$ \\
Xiaoye Wang$^{1}$ \quad
Kailun Yang$^{3}$ \quad
Danda Pani Paudel$^{1}$ \quad
Luc Van Gool$^{1}$ \quad
Kunyu Peng$^{2,\dag}$
\thanks{$^{1}$INSAIT, Sofia University ``St. Kliment Ohridski'', Bulgaria} \\
\thanks{$^{2}$Karlsruhe Institute of Technology, Germany} \\
\thanks{$^{3}$Hunan University, China}\\
\thanks{*Equal contribution}\\ 
\thanks{$\dag$Corresponding author: kunyu.peng@kit.edu}
}
\IEEEaftertitletext{
\begin{minipage}{\textwidth}
\centering
\vspace{-5em}
\vskip-5ex
\includegraphics[width=\textwidth]{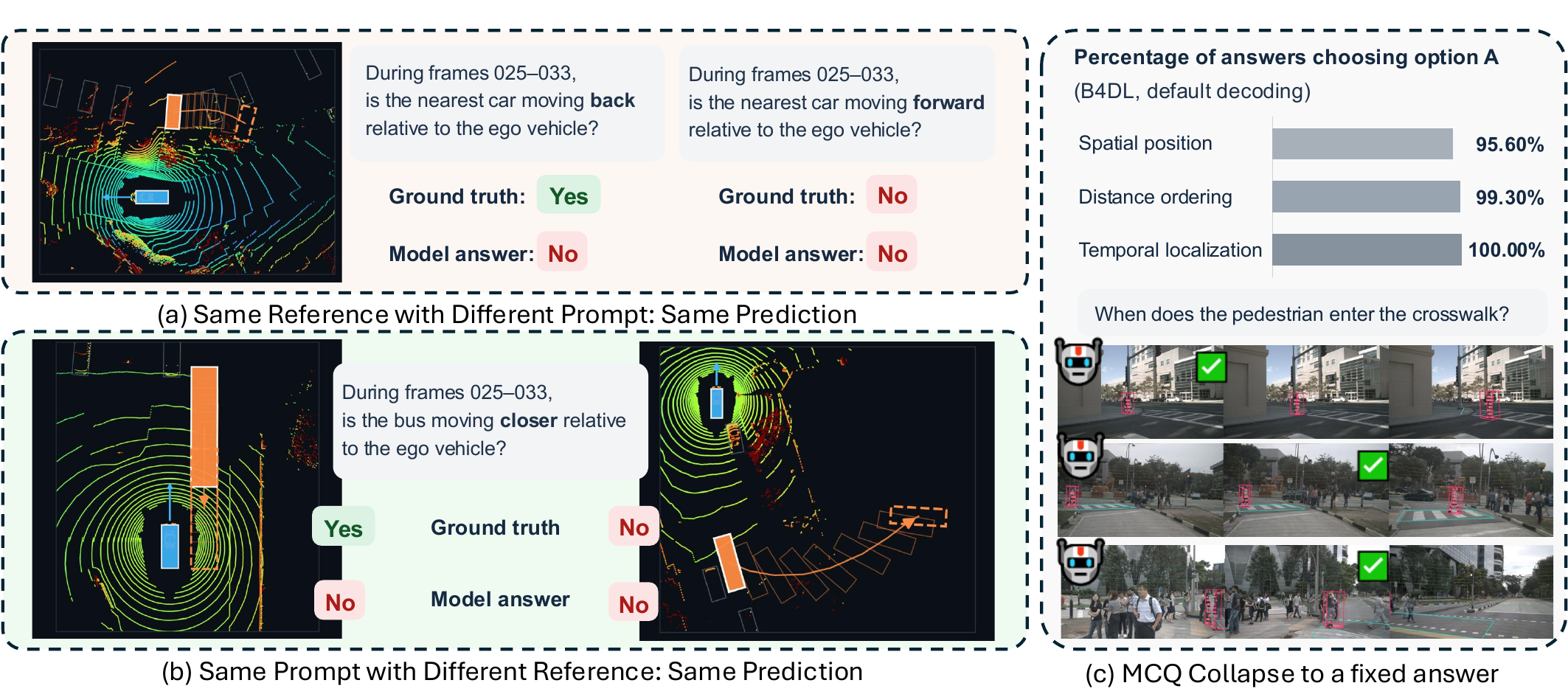}
\captionof{figure}{\textbf{Response collapse in LiDAR QA.} (a) Different motion queries about the same car receive ``No'' despite opposite reference answers. (b) The same query receives ``No'' in two scenes requiring opposite answers. (c) Option-A selection rates for B4DL (MCQ stands for Multi-choice Question).}
\label{fig:teaser}
\end{minipage}
\par\medskip
}
\begin{document}
\maketitle
\begin{abstract}
Recent 4D LiDAR language models aim to reason about objects and their evolving spatial relationships. Yet, in our evaluation, always selecting the same option nearly matches the multiple-choice accuracy of two B4DL-derived configurations. We introduce \bench{}, a geometry-referenced benchmark and diagnostic protocol with 10,000 questions across 150 nuScenes scenes. It covers object existence, ego-relative position, distance ordering, relative motion, and temporal localization, with explicit rules for selecting objects, comparing times, and determining reference answers. Our protocol combines fixed-answer and candidate-content controls, cross-scene pairs with identical prompts but opposite reference answers, and relation-specific recall. Analysis of 100,000 recorded responses reveals failures hidden by aggregate accuracy. Candidate duration alone makes temporal answers predictable without observing LiDAR. On paired questions, the models frequently give the same answer to scenes requiring opposite answers. Relation-specific analysis further shows that both configurations miss every positive lateral-motion case across all tested conditions. Temporal-shuffle contrastive decoding provides little net improvement, as repairs are largely offset by new errors and the main failures persist. These results show that evaluating spatio-temporal reasoning requires testing whether models distinguish the queried physical relationships, rather than relying on individual-answer accuracy alone. The source code, checkpoints, and data are released at this \href{https://github.com/Awesome4D/4DMLLM_Hallucination_Bench}{repository}.
\end{abstract}

\section{Introduction}
\label{sec:intro}
Autonomous driving requires understanding where surrounding objects are and how they move relative to the ego vehicle~\cite{caesar2020nuscenes}. Decisions about yielding, passing, and maintaining separation depend on which obstacle is closest, whether another vehicle is approaching, and when an event occurs~\cite{sima2024drivelm}. Recent methods connect LiDAR observations with language to support scene understanding~\cite{hess2022lidarclip,yang2025lidar}, while question-answering benchmarks assess these capabilities. Such judgments require more than recognizing objects. Determining whether a car is approaching requires following the same car and comparing its distance to the ego vehicle over the specified interval.

B4DL~\cite{choi2025b4dl} extends LiDAR-language modeling to point-cloud sequences, providing a benchmark and an accompanying model for spatial and temporal question answering. However, average accuracy can conceal systematic errors. Figure~\ref{fig:teaser}(a) shows two questions about the same car over the same interval, asking whether it is moving back or forward relative to the ego vehicle. The reference answers are ``Yes'' and ``No'', but the model answers ``No'' to both. It answers one question correctly while missing the motion identified by the reference. Panel~(b) holds the question fixed and changes the scene: the reference answer changes, but the model again answers ``No'' to both. In multiple-choice questions (MCQ), always selecting ``A'' nearly matches the accuracy of the two B4DL-derived configurations evaluated here. Panel~(c) shows this fixed-option behavior across spatial and temporal tasks.

Evaluation must therefore examine both answers that are predictable without the scene and relations that a model fails to distinguish. Existing driving benchmarks cover object-centric questions~\cite{qian2024nuscenesqa}, perception and planning~\cite{sima2024drivelm}, spatial measurements~\cite{tian2025spatialqa}, and cross-view temporal reasoning~\cite{vo2026drivespatial}. In static 3D question answering, text-only models can match or exceed 3D models on SQA3D~\cite{ma2026real3dqa}. 
Testing for answer biases is therefore only a first step, as it reveals whether predictions can be driven by textual or distributional cues without observing the scene. We further use paired questions and relation-specific recall to test whether the model actually distinguishes the physical relations that determine the correct answer.

\begin{figure}[t]
\centering
\includegraphics[width=0.95\linewidth]{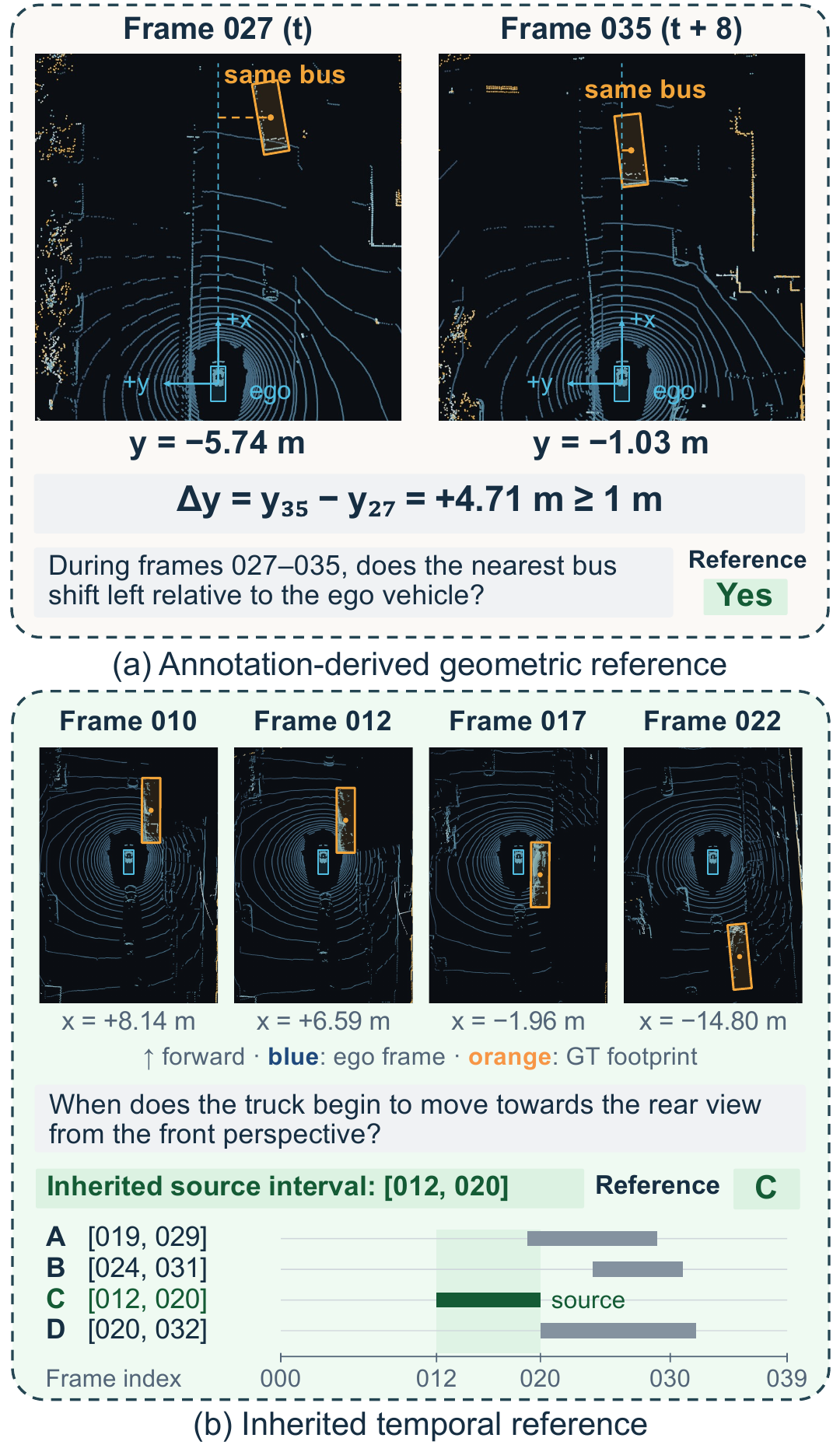}
\vspace{-1ex}
\caption{\textbf{Question and reference construction.} (a) Tracking the same bus gives its ego-relative lateral change. (b) A source event interval and three distractors form a temporal-localization question.}
\label{fig:qa_construction}
\end{figure}

We introduce \bench{}, a benchmark and diagnostic protocol with 10,000 questions across 150 nuScenes scenes~\cite{caesar2020nuscenes}. Its five task families cover existence, position, distance, motion, and event timing. Each family specifies the queried entities, relevant times, coordinate frame, and answer rule. The first four derive answers from nuScenes annotations. Temporal localization retains B4DL's event intervals and adds three distractors (Figure~\ref{fig:qa_construction}). Table~\ref{tab:benchmark_comparison} compares this combination of sequence question answering and diagnostic evaluation with prior benchmarks.

We evaluate two B4DL-derived configurations under five inference settings, obtaining 100,000 responses to the same 10,000 questions. Three questions guide the study:
\begin{enumerate}
\item[RQ1.] \textit{What can be answered without observing LiDAR?} We measure fixed-answer performance and predict answers from questions and candidate text using statistics from other scenes.
\item[RQ2.] \textit{Can models distinguish opposite physical relations?} We measure joint correctness on exact-prompt cross-scene pairs and positive recall for individual relations.
\item[RQ3.] \textit{Can inference-time interventions correct these errors?} We evaluate anti-position prompting and Temporal-Shuffle Contrastive Decoding (TSCD), counting repaired and newly introduced errors separately.
\end{enumerate}
Our contributions are threefold:
\begin{itemize}
\item A 10,000-question LiDAR benchmark with five task families, explicit reference-answer rules, and executable integrity checks.
\item A diagnostic protocol combining answer-bias controls, exact-prompt pairs, relation-specific recall, and paired correction analysis.
\item An analysis of 100,000 responses that identifies fixed-answer behavior, systematic relation omissions, and the limited benefit of the tested inference-time interventions.
\end{itemize}

\section{Related Work}
\label{sec:related}
\noindent\textbf{Driving-scene question answering.}
NuScenes-QA generates questions about object existence, counts, states, and spatial relations from annotated scene graphs~\cite{qian2024nuscenesqa}. DriveLM links perception, prediction, and planning through graph-structured questions~\cite{sima2024drivelm}. LingoQA evaluates scene descriptions, anticipation, and action justification in driving videos~\cite{marcu2024lingoqa}, while DriveGPT4 combines video question answering and action explanations with vehicle-control prediction~\cite{xu2024drivegpt4}. NuScenes-SpatialQA focuses on qualitative spatial relations and quantitative measurements~\cite{tian2025spatialqa}, and DriveSpatial examines cross-view and temporal reasoning~\cite{vo2026drivespatial}.

\noindent\textbf{LiDAR-language modeling.}
LidarCLIP aligns automotive point clouds with CLIP embeddings for retrieval and cross-modal applications~\cite{hess2022lidarclip}. LiDAR-LLM connects point clouds to captioning, question answering, and planning through a view-aware transformer and staged training~\cite{yang2025lidar}. B4DL extends this setting to sequences, with spatial reasoning, temporal localization, and scene-dynamics tasks~\cite{choi2025b4dl}. \bench{} builds on this setting with explicit geometric reference rules and tests whether answers distinguish the corresponding relations, as shown in Table~\ref{tab:benchmark_comparison}.

\begin{table*}[t]
\centering
\caption{Driving-scene language benchmarks: sensor input, temporal coverage, and diagnostic evaluation. Dataset characteristics follow Table~1 of B4DL~\cite{choi2025b4dl}. The diagnostic columns indicate reported evaluations.}
\label{tab:benchmark_comparison}
\footnotesize
\setlength{\tabcolsep}{2.5pt}
\renewcommand{\arraystretch}{1.15}
\begin{tabular*}{\textwidth}{@{\extracolsep{\fill}}l*{9}{c}@{}}
\toprule
\multicolumn{7}{c}{Dataset characteristics} & \multicolumn{3}{c}{Diagnostic evaluation}\\
\cmidrule(lr){1-7}\cmidrule(l){8-10}
Dataset & Input & \shortstack{LiDAR-\\specific} & \shortstack{$360^\circ$\\coverage} & \shortstack{Multi-\\frame} & Sequence & \shortstack{Sequence\\annotation} & \shortstack{Answer-bias\\controls} & \shortstack{Exact-prompt\\pairs} & \shortstack{Relation\\recall}\\
\midrule
DriveLM~\cite{sima2024drivelm} & Camera & $\times$ & $\checkmark$ & $\checkmark$ & $\times$ & $\times$ & $\times$ & $\times$ & $\times$ \\
\addlinespace[2pt]
LingoQA~\cite{marcu2024lingoqa} & Camera & $\times$ & $\times$ & $\checkmark$ & $\checkmark$ & $\checkmark$ & $\checkmark$ & $\times$ & $\times$ \\
\addlinespace[2pt]
DriveGPT4~\cite{xu2024drivegpt4} & Camera & $\times$ & $\times$ & $\checkmark$ & $\checkmark$ & $\checkmark$ & $\times$ & $\times$ & $\times$ \\
\addlinespace[2pt]
nuScenes-QA~\cite{qian2024nuscenesqa} & \shortstack{Camera\\+ LiDAR} & $\checkmark$ & $\checkmark$ & $\checkmark$ & $\checkmark$ & $\times$ & $\checkmark$ & $\times$ & $\times$ \\
\addlinespace[2pt]
LiDARLLM~\cite{yang2025lidar} & LiDAR & $\checkmark$ & $\checkmark$ & $\times$ & $\times$ & $\times$ & $\times$ & $\times$ & $\times$ \\
\addlinespace[2pt]
B4DL~\cite{choi2025b4dl} & LiDAR & $\checkmark$ & $\checkmark$ & $\checkmark$ & $\checkmark$ & $\checkmark$ & $\times$ & $\times$ & $\times$ \\
\addlinespace[2pt]
\midrule
\textbf{\bench{} (ours)} & LiDAR & $\checkmark$ & $\checkmark$ & $\checkmark$ & $\checkmark$ & $\checkmark$ & $\checkmark$ & $\checkmark$ & $\checkmark$\\
\bottomrule
\end{tabular*}
\end{table*}

\noindent\textbf{Answer biases and relational evaluation.}
VQA v2 pairs identical questions with complementary images requiring different answers~\cite{goyal2017making}, while VQA-CP changes question-conditioned answer distributions between training and testing~\cite{agrawal2018priors}. NuScenes-QA and LingoQA include question-only or text-only baselines~\cite{qian2024nuscenesqa,marcu2024lingoqa}. Real-3DQA studies textual shortcuts in 3D question answering and filters questions readily answered without 3D input~\cite{ma2026real3dqa}. Multiple-choice studies also identify option-order sensitivity and weaknesses in binding answer symbols to candidates~\cite{pezeshkpour2024order,xue2024symbol}. Our protocol measures fixed-answer behavior and candidate-content predictability separately, then tests joint correctness under identical complete prompts and opposite reference answers.

\noindent\textbf{Diagnostic benchmarks and mitigation.}
Winoground and ARO test compositional, attribute, and relation distinctions~\cite{thrush2022winoground,yuksekgonul2023aro}. Beacon3D examines object-centric descriptions and agreement between grounding and question answering~\cite{huang2025beacon3d}. 4D-Bench and Spatial4D-Bench address dynamic objects and spatio-temporal reasoning~\cite{zhu2025four,wang2026spatial}. CHAIR, POPE, and AMBER assess unsupported object and relational claims~\cite{rohrbach2018object,li2023evaluating,wang2023amber}, while VidHalluc and EventHallusion study temporal and event-level errors~\cite{li2025vidhalluc,zhang2024eventhallusion}. We measure both false assertions and missed true relations in LiDAR sequences. For correction, visual and temporal contrastive decoding methods compare predictions from original and perturbed inputs~\cite{leng2024vcd,zhang2024eventhallusion}. We adapt temporal shuffling to LiDAR features and measure whether its repairs outweigh the errors it introduces.

\section{The LiDAR-Hallu Benchmark}
\label{sec:benchmark}
\begin{figure}[t]
\centering
\includegraphics[width=0.95\linewidth]{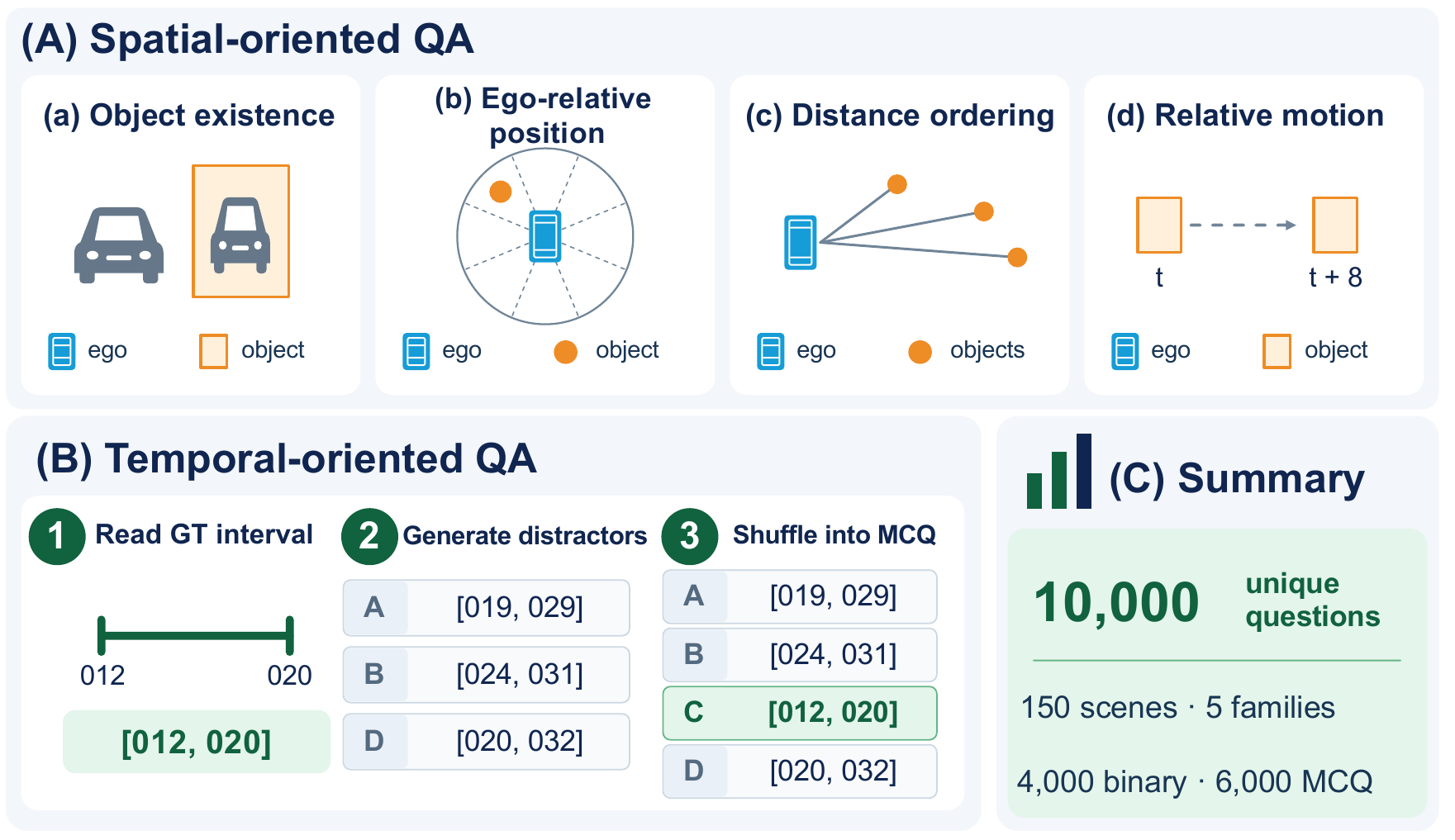}
\vspace{-1ex}
\caption{\textbf{Benchmark overview.} Four task families use object annotations and ego poses. Temporal questions use source event intervals and sampled distractors.}
\label{fig:pipeline}
\end{figure}

\bench{} contains 10,000 unique questions across 150 nuScenes scenes~\cite{caesar2020nuscenes}, with 2,000 per family (Table~\ref{tab:benchmark}). Object existence, ego-relative position, distance ordering, and relative motion derive answers from object annotations and ego poses. Temporal localization retains B4DL's event questions and reference intervals~\cite{choi2025b4dl}. Figure~\ref{fig:pipeline} summarizes their construction.

\subsection{Reference answers and coordinates}
Each geometry-derived question specifies entities $\mathcal E_q$, frames $\mathcal T_q$, a coordinate frame $\mathcal F_q$, and an answer rule $\phi_q$:
\begin{equation}
\begin{aligned}
\mathcal C_q &= (\mathcal E_q,\mathcal T_q,\mathcal F_q,\phi_q),\\
y_q &= \phi_q(\mathcal A;\mathcal E_q,\mathcal T_q,\mathcal F_q),
\end{aligned}
\label{eq:contract}
\end{equation}
where $\mathcal A$ denotes the nuScenes validation annotations. The rule evaluates class membership, a spatial or motion relation, or distance ordering.

For global object center $\mathbf p_{o,t}$ and ego pose $(\mathbf R_t,\mathbf t_t)$, we compute
\begin{equation}
\begin{aligned}
\mathbf p^e_{o,t} &= \mathbf R_t^\top(\mathbf p_{o,t}-\mathbf t_t),\\
d_{o,t} &= \sqrt{(x^e_{o,t})^2+(y^e_{o,t})^2}.
\end{aligned}
\label{eq:ego}
\end{equation}
The ego axes point forward ($x$), left ($y$), and up ($z$). Distance is measured from the ego origin to the object center in the ground plane. We use ten classes: car, truck, bus, pedestrian, bicycle, motorcycle, trailer, construction vehicle, barrier, and traffic cone. Annotations reporting zero LiDAR returns are excluded. The index $t$ follows the annotated keyframes at 2~Hz~\cite{caesar2020nuscenes}.

\begin{table}[t]
\centering\small
\caption{Benchmark composition and reference answers. MCQ denotes a four-choice question.}
\label{tab:benchmark}
\begin{tabular}{@{}lrrl@{}}
\toprule
Family & Binary & MCQ & Reference\\
\midrule
Existence & 1,000 & 1,000 & Class membership\\
Position & 1,000 & 1,000 & Ego-frame sector\\
Distance & 0 & 2,000 & Nearest class\\
Motion & 2,000 & 0 & Same-instance deltas\\
Temporal & 0 & 2,000 & Source interval\\
\midrule
Total & 4,000 & 6,000 & 150 scenes\\
\bottomrule
\end{tabular}
\end{table}

\subsection{Question construction}
\noindent\textit{Object existence.}
Binary questions ask whether a class is present at a specified keyframe, with 500 positive and 500 negative examples. Each of the 1,000 multiple-choice questions contains one present class and three absent classes.

\noindent\textit{Ego-relative position.}
We assign the nearest instance of the queried class to one of eight $45^\circ$ bearing sectors: front, front-left, left, back-left, back, back-right, right, and front-right. Binary questions balance correct and incorrect proposed sectors. Multiple-choice questions contain the reference sector and three alternatives.

\noindent\textit{Distance ordering.}
Classes are ranked by the distance of their nearest instance. The closest class is the reference, with three other present classes as distractors. Near-tied distances are retained.

\noindent\textit{Relative motion.}
We select the nearest instance of the queried class at keyframe $t$ and follow it to $t+8$, approximately 4~s later, requiring annotations at both endpoints (Figure~\ref{fig:qa_construction}(a)). Using each endpoint's ego coordinates, we compute $\Delta d=d_{o,t+8}-d_{o,t}$ and $\Delta y=y^e_{o,t+8}-y^e_{o,t}$. The prompts use five qualitative relations, with positive labels defined by
\begin{align}
\text{closer/farther}:&\quad
\Delta d\leq-2\,\mathrm{m}\;/\;\Delta d\geq2\,\mathrm{m}, \nonumber\\
\text{left/right}:&\quad
\Delta y\geq1\,\mathrm{m}\;/\;\Delta y\leq-1\,\mathrm{m}, \nonumber\\
\text{similar}:&\quad
|\Delta d|<1\,\mathrm{m}\ \land\ |\Delta y|<0.5\,\mathrm{m}.
\label{eq:motion}
\end{align}
Each predicate is evaluated independently, receiving ``Yes'' when its condition holds and ``No'' otherwise. Sub-threshold directional changes remain negative candidates. We sample 1,000 positive and 1,000 negative questions across the five predicates. 
\noindent\textit{Temporal localization.}
Each source event interval is paired with three distinct, in-bounds distractors (Figure~\ref{fig:qa_construction}(b)). A distractor $[s,e]$ has frame-index span $\ell=e-s\in\{5,\ldots,19\}$, covering 6--20 frames including both endpoints. Its intersection with the reference, divided by their enclosing span, must be at most 0.2. All spans use endpoint differences. The four candidates are randomly ordered.

\subsection{Answer distribution and records}
The 6,000 multiple-choice questions have A/B/C/D reference counts of 1,474/1,572/1,453/1,501. Among the temporal references, 242 spans fall outside the distractor range of 5--19. We measure the resulting candidate-content predictability in Sec.~\ref{sec:protocol}.

Records store prompts, candidates, references, scene/frame identifiers, and motion deltas. All prompt variants share the same questions and labels. Integrity checks verify record correspondence, candidate uniqueness, motion labels, interval bounds, and temporal-shuffle permutations using the recorded feature-sequence lengths of 39--41 steps.

\subsection{Exact-prompt contrast pairs}
\label{sec:contrastset}
Within each binary family, we group questions by complete prompt and feature-sequence length. Deterministic maximum bipartite matching pairs opposite reference answers from different scenes, using each question at most once. Selection follows a fixed scene and question-index order without using model predictions.

B4DL has 348 pairs: 104 existence, 41 position, and 203 motion, covering 696 questions and 140 scenes. B4DL-Ego additionally requires identical ego-motion text, producing a separate set of 38 pairs across 56 scenes. These configuration-specific sets remain fixed across inference settings and are evaluated using joint correctness in Sec.~\ref{sec:protocol}.

\section{A Diagnostic Evaluation Protocol}
\label{sec:protocol}
We test answer biases, discrimination between physical relations, and the net benefit of inference-time correction. Accuracies and rates in the equations are proportions.

\subsection{Answer parsing and fixed-answer controls}
We map each response $r_i$ to a label $\hat y_i=\mathcal P(r_i)$. The parser accepts explicit Yes/No answers, A--D option labels or answer declarations, and unique exact matches to candidate text. A leading explicit label takes precedence over the explanation. Invalid responses count as incorrect, and valid-response rates are reported separately.

The fixed-answer controls select A for multiple-choice questions and No for binary questions. On $N$ multiple-choice questions, let $A$ denote model accuracy, $\pi_A$ the proportion of reference answers labeled A, and $s_A$ the model's A-selection rate. If $R_A$ counts correct non-A predictions and $D_A$ counts reference-A questions answered otherwise, then
\begin{equation}
A-\pi_A=\frac{R_A-D_A}{N},
\qquad |A-\pi_A|\leq1-s_A.
\label{eq:position}
\end{equation}
Thus only departures from A can change the score relative to that fixed answer.

\subsection{Candidate-content predictability}
For each scored scene $s$, we estimate answer frequencies from all other scenes within the same task. The controls use reference labels to estimate these frequencies and only question and candidate text to predict answers. Multiple-choice features are class names for existence and distance, sectors conditioned on the queried class for position, and frame-index spans for temporal localization. The temporal control uses neither event descriptions nor absolute endpoints.

Let $n^+_{-s,c}(v)$ and $n^-_{-s,c}(v)$ count reference and distractor occurrences of feature $v$, excluding scene $s$. Context $c$ is the queried class for position and the task alone for existence, distance, and temporal localization. For candidate feature $v_k$, we use
\begin{equation}
\rho_{-s,c}(v)=\frac{n^+_{-s,c}(v)+1}{n^-_{-s,c}(v)+1},
\qquad \hat y=\arg\max_k\rho_{-s,c}(v_k).
\label{eq:content}
\end{equation}
Here $k$ ranges over option labels A--D, with ties resolved by candidate order. Binary controls predict the majority label conditioned on class, class--sector, or class--action, breaking ties as No. These leave-one-scene-out controls measure predictability within the benchmark. We also evaluate the models on questions the controls answer incorrectly.

\subsection{Joint correctness under identical prompts}
For the $M$ pairs in Sec.~\ref{sec:contrastset}, let $(\hat y_i,\hat y'_i)$ denote predictions and $(y_i,y'_i)$ their opposite binary references. A pair is correct only when both answers are correct:
\begin{equation}
J=\frac{1}{M}\sum_{i=1}^{M}
\ind[\hat y_i=y_i\ \land\ \hat y'_i=y'_i].
\label{eq:joint}
\end{equation}
We report $J$, individual-question accuracy, and the same-answer rate. A deterministic Yes/No predictor using only the shared prompt achieves 50\% individual accuracy and zero joint correctness. Under independent prompt-only sampling, expected joint success for a pair is $p(1-p)\leq1/4$, where $p$ is the probability of answering Yes. These give deterministic and sampling references for interpreting the paired results.

\subsection{Relation recall and correction accounting}
For each binary relation, we report positive and negative counts, true positive recall (TPR), and false-positive rate (FPR). On a balanced subset with all responses valid,
\begin{equation}
A=\tfrac12(\mathrm{TPR}+1-\mathrm{FPR}).
\label{eq:binary}
\end{equation}
Reporting both rates shows whether higher accuracy comes from recognizing more true relations or rejecting more false ones.

We compare inference settings on the same $N$ questions. If $R$ counts wrong-to-correct repairs and $D$ counts correct-to-wrong regressions, then
\begin{equation}
\Delta A=(R-D)/N.
\label{eq:paired}
\end{equation}
Wrong-to-wrong changes are recorded separately. For item accuracy and accuracy differences, 95\% confidence intervals use 10,000 whole-scene bootstrap samples with seed 20260904. Each draw recomputes question-weighted scores and keeps predictions from compared settings paired. Predictions and fitted content-control outputs remain fixed during resampling.

\section{Experiments}
\label{sec:experiments}
\subsection{Models, training, and inference}
\begin{figure}[t]
    \centering
    \includegraphics[width=0.95\linewidth]{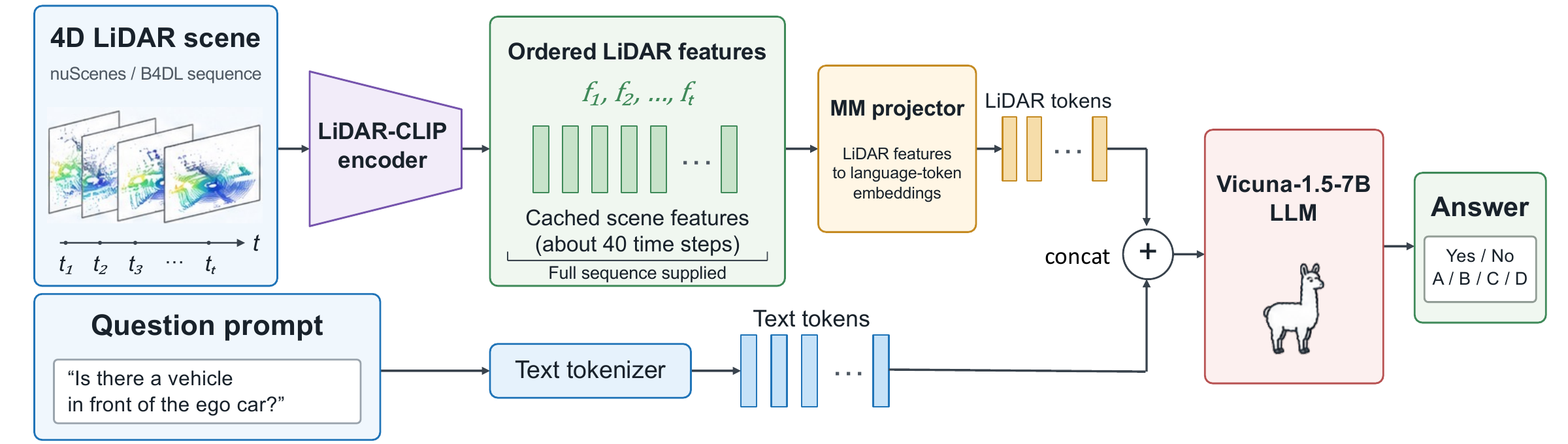}
    \caption{\textbf{Evaluated model architecture.} Projected LiDAR-CLIP features and question tokens form the input to Vicuna-1.5-7B. B4DL-Ego also includes ego-motion text.}
    \label{fig:model_architecture}
\end{figure}

\noindent\textbf{Architecture and inputs.}
We evaluate B4DL and B4DL-Ego, two B4DL-derived checkpoints instruction-tuned separately on B4DL's nuScenes-based data~\cite{choi2025b4dl}. Both use a learned projector to map cached LiDAR-CLIP features~\cite{hess2022lidarclip} into Vicuna-1.5-7B token embeddings, as shown in Figure~\ref{fig:model_architecture}. 
B4DL takes LiDAR features and question text as input, while B4DL-Ego also receives ego-motion text. All questions, including single-frame queries, use the full scene sequence of 39--41 feature steps in temporal order. 
Ego-motion text summarizes ego speed, displacement direction, heading change, and acceleration from local pose changes at two selected endpoints. The enclosing span is determined by 
all frame intervals in the prompt, including those in the answer choices. 
Single-frame queries use 
a five-frame window adjusted at scene boundaries, while prompts without frame indices use the scene endpoints. Neither model receives object annotations or reference answers at inference.

\noindent\textbf{Training configuration.}
The archived training configuration specifies one epoch of projector alignment at a learning rate of $10^{-3}$, followed by two epochs of Vicuna instruction tuning at $10^{-4}$ with the projector frozen. 
Instruction tuning uses LoRA with rank 64 and $\alpha_{\mathrm{LoRA}}=128$, combining B4DL's stage-2 and stage-3 records with ego-motion text added for B4DL-Ego. Alignment utilizes a per-GPU batch size of 16 with eight gradient-accumulation steps, while instruction tuning applies a per-GPU batch size of eight with 16 accumulation steps. 
Both stages use fixed cached features, FP16, cosine learning-rate decay, 3\% warmup, zero weight decay, and a 2,048-token context limit.

\noindent\textbf{Inference.}
We evaluate each model 
with default decoding, anti-position prompting, and TSCD at $\alpha\in\{0.5,1,1.5\}$. Standard prompts request Yes/No or an A--D label. Anti-position prompting adds ``Do not favor any option based on its position; evaluate all options equally before selecting your answer.'' to MCQ 
instructions only, leaving candidate order and binary prompts unchanged. All settings sample at temperature 0.05, one beam, and at most 32 new tokens. We generate one response per question and setting, 
yielding $100{,}000$ responses from the two models on the same $10{,}000$ questions. Invalid responses are counted as incorrect under the scoring protocol in Sec.~\ref{sec:protocol}. 

\subsection{Temporal-shuffle contrastive decoding}
\label{sec:tscd}
Following visual and temporal contrastive decoding~\cite{leng2024vcd,zhang2024eventhallusion}, TSCD contrasts the ordered feature sequence $H$ with a temporal permutation $\widetilde H$, drawn once per question. Both branches use the same checkpoint, question, and ego-motion text when present. At decoding step $j$, they share the generated answer prefix but maintain separate key/value caches. The combined logits are
\begin{equation}
\mathbf z_j^\alpha=\mathbf z_j^c+\alpha(\mathbf z_j^c-\mathbf z_j^s),
\label{eq:tscd}
\end{equation}
where $c$ and $s$ denote the clean and shuffled branches, respectively. For $\alpha>0$, the contrastive term increases the logits of tokens whose clean logits exceed shuffled logits. 

The permutation and sampling seeds are $42+i$ and $1234+i$, respectively, where $i$ is the zero-based question index within each task family. TSCD uses a token-wise decoding loop without an adaptive token-plausibility mask. 
Default and anti-position runs use \texttt{model.generate} without an explicit sampling seed.

\subsection{RQ1: Fixed answers and candidate cues}
\label{sec:results_priors}
\begin{table}[t]
\centering\small
\caption{Accuracy (\%) by task and answer format. B: binary; M: four-choice. Fixed answers are No for B and A for M. Content controls use statistics from other scenes.}
\label{tab:main}
\begin{tabular}{@{}lrrrr@{}}
\toprule
Task & Fixed & \nm{} & \mt{} & Content\\
\midrule
Existence (B) & 50.00 & 66.30 & 67.20 & 72.10\\
Existence (M) & 25.00 & 26.70 & 26.30 & 66.10\\
Position (B) & 50.00 & 45.70 & 44.50 & 54.90\\
Position (M) & 25.50 & 26.10 & 27.00 & 29.10\\
Motion (B) & 50.00 & 52.40 & 50.00 & 59.05\\
Distance (M) & 24.30 & 24.30 & 24.40 & 52.70\\
Temporal (M) & 24.15 & 24.15 & 24.15 & 69.30\\
\midrule
Overall & 34.74 & 36.65 & 36.21 & 58.43\\
\bottomrule
\end{tabular}
\end{table}

\noindent\textbf{Multiple-choice accuracy largely follows a fixed answer.}
Under default decoding, both models remain within 1.70 percentage points of constant A across the four MCQ tasks as in Table~\ref{tab:main}. Both select A in more than 96\% of distance-ordering and temporal-localization responses, constraining their accuracy differences from constant A, per Eq.~\eqref{eq:position}.


The margins are larger on binary existence questions, \textit{i.e.,} B4DL and B4DL-Ego achieve 66.30\% and 67.20\% accuracy, respectively, compared with 50.00\% for constant No. B4DL-Ego produces more varied spatial-position answers, but achieves 27.00\% , compared with 26.10\% for B4DL. 

\noindent\textbf{Candidate contents remain informative.}
The scene-held-out controls, in Table~\ref{tab:main}, are fitted using reference labels from the other benchmark scenes as in Sec.~\ref{sec:protocol}. Candidate class names yield 66.10\% accuracy on existence MCQs and 52.70\% on distance ordering. Although every distance candidate is present in the scene, its class remains informative about whether it is the nearest object. 

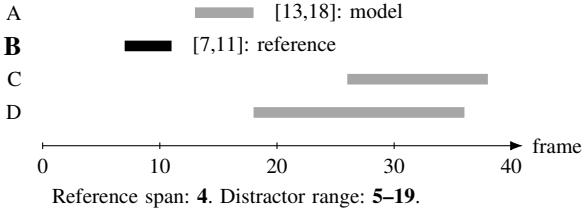
\begin{figure}[t]
\centering
\begin{tikzpicture}[x=0.155cm,y=0.44cm,font=\footnotesize,>=Latex]
\node[anchor=west,align=left,text width=7.5cm] at (0,5.65)
{\textit{During which frames do pedestrians appear on the sidewalks?}};
\draw[->] (0,0)--(41,0) node[right]{frame};
\foreach \x in {0,10,20,30,40}{\draw (\x,0.10)--(\x,-0.10) node[below]{\x};}
\node[anchor=east] at (-1,4){A};\draw[line width=4pt,black!35] (13,4)--(18,4);\node[anchor=west] at (19,4){[13,18]: model};
\node[anchor=east,font=\bfseries] at (-1,3){B};\draw[line width=4pt] (7,3)--(11,3);\node[anchor=west] at (12,3){[7,11]: reference};
\node[anchor=east] at (-1,2){C};\draw[line width=4pt,black!35] (26,2)--(38,2);
\node[anchor=east] at (-1,1){D};\draw[line width=4pt,black!35] (18,1)--(36,1);
\node[anchor=north west,align=left,text width=7.5cm] at (0,-1.0)
{Reference span: \textbf{4}. Distractor range: \textbf{5--19}.};
\end{tikzpicture}
\caption{\textbf{Temporal candidate spans.} The span-only control selects B, while B4DL selects A under default decoding (record 5, scene 006599465).
}
\label{fig:temporalcase}
\end{figure}

Temporal localization shows the clearest candidate cue. The span-only control achieves 69.30\% accuracy without event descriptions or LiDAR input, 
whereas both models always select A and score 24.15\%. 
In Figure~\ref{fig:temporalcase}, the reference intervel $[7,11]$ has span 4, below the permitted distractor range of 5--19, 
which length reveals the answer. The control selects it, whereas B4DL stays with A. 

\subsection{RQ2: Distinguishing physical relations}
\label{sec:results_joint}
\begin{table}[t]
\centering\small
\caption{B4DL on 348 exact-prompt pairs with opposite reference answers. The prompt-only row gives the deterministic reference.}
\label{tab:joint}
\begin{tabular}{@{}lrrr@{}}
\toprule
Condition & Item acc. (\%) & Both correct & Joint (\%)\\
\midrule
Prompt-only null & 50.00 & 0/348 & 0.00\\
\midrule
Baseline & 53.30 & 43/348 & 12.36\\
Anti-position & 51.87 & 39/348 & 11.21\\
TSCD $\alpha=0.5$ & 52.16 & 42/348 & 12.07\\
TSCD $\alpha=1$ & 52.30 & 42/348 & 12.07\\
TSCD $\alpha=1.5$ & 52.01 & 40/348 & 11.49\\
\bottomrule
\end{tabular}
\end{table}

\noindent\textbf{Individual accuracy hides failed contrasts.}
On 348 exact-prompt pairs with opposite reference answers, B4DL achieves 53.30\% accuracy on individual questions but answers both questions correctly in only 43 pairs as in Table~\ref{tab:joint}. Most pairs receive identical 
answers, and only one of the 41 spatial-postion pairs is jointly correct. A deterministic prompt-only predictor has zero joint correctness on these pairs, although sampling can produce nonzero joint correctness without scene information. 

Anti-position prompting and TSCD improves yield 39 and 40--42 jointly correct B4DL pairs, respectively, compared with 43 under default decoding. 
On the separate 38-pair B4DL-Ego set, four pairs are jointly correct with TSCD, compared with two under default decoding, 
while most pairs remain unresolved. 
Each configuration is compared across inference settings on its own fixed pair set.

\begin{figure*}[t]
\centering
\includegraphics[width=0.94\linewidth]{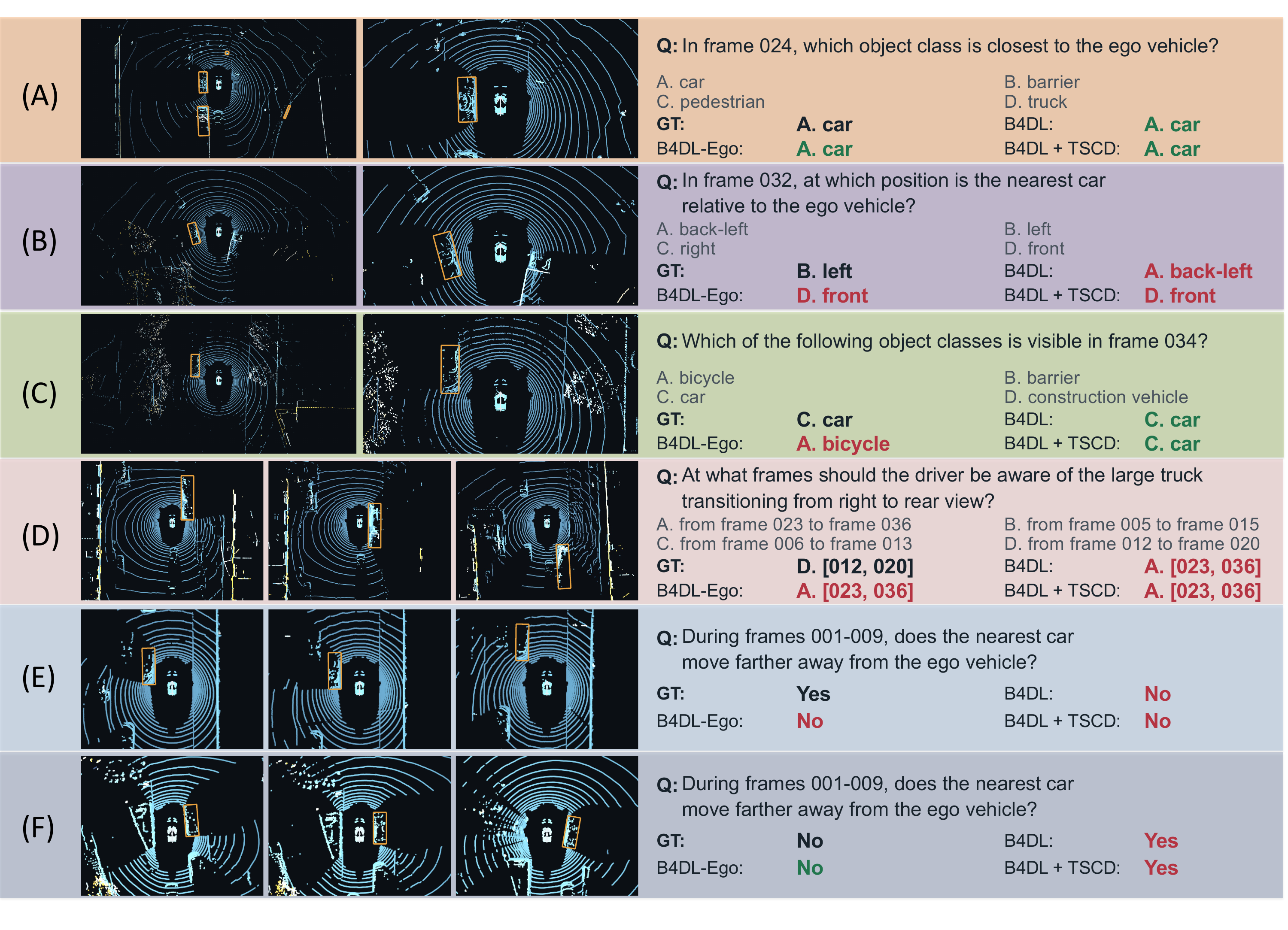}
\vspace{-1ex}
\caption{\textbf{Qualitative results on \bench{}.} Examples cover distance, position, existence, temporal localization, and paired motion questions. Predictions compare default decoding with TSCD.}
\label{fig:qualitative}
\vspace{-1em}
\end{figure*}

\noindent\textbf{Errors concentrate in particular relations.}
\label{sec:results_predicates}
\begin{table}[t]
\centering\footnotesize
\caption{Motion labels, positive recall, and accuracy (\%) under default decoding.}
\label{tab:motion}
\begin{tabular}{@{}llrrrr@{}}
\toprule
& & \multicolumn{2}{c}{Positive recall} & \multicolumn{2}{c}{Accuracy}\\
\cmidrule(lr){3-4}\cmidrule(l){5-6}
Relation & Yes / No & \nm{} & \mt{} & \nm{} & \mt{}\\
\midrule
Closer & 206 / 209 & 4.37 & 7.77 & 45.78 & 44.58\\
Farther & 347 / 141 & 30.26 & 7.20 & 43.44 & 33.81\\
Left & 145 / 199 & 0.00 & 0.00 & 57.85 & 57.85\\
Right & 157 / 221 & 0.00 & 0.00 & 58.47 & 58.47\\
Similar & 145 / 230 & 55.86 & 61.38 & 60.27 & 61.33\\
\bottomrule
\end{tabular}
\end{table}

Table~\ref{tab:motion} explains why motion accuracy alone gives an incomplete picture. Both models answer No to every left- and right-motion question under all evaluated settings. 
All 302 positive lateral-motion cases are missed. The corresponding accuracies of 57.85\% and 58.47\% equal the proportions of negative examples, with zero positive recall for both predicates. 

Recall also varies between distance-change predicates, \textit{i.e.,} closer recall is below 8\% for both models, whereas farther recall is 30.26\% for B4DL and 7.20\% for B4DL-Ego. Recall for the predefined Similar predicate is higher, at 55.86\% and 61.38\%, respectively,
indicating relation-specific errors.

\begin{table}[t]
\centering\small
\caption{Positive recall (TPR) and false-positive rate (FPR), in \%, under default decoding.}
\label{tab:binary}
\begin{tabular}{@{}lrrrr@{}}
\toprule
& \multicolumn{2}{c}{TPR $\uparrow$} & \multicolumn{2}{c}{FPR $\downarrow$}\\
\cmidrule(lr){2-3}\cmidrule(l){4-5}
Family & \nm{} & \mt{} & \nm{} & \mt{}\\
\midrule
Existence & 49.80 & 40.60 & 17.20 & 6.20\\
Position & 69.80 & 63.80 & 78.40 & 74.80\\
Motion & 19.50 & 13.00 & 14.70 & 13.00\\
\bottomrule
\end{tabular}
\vspace{-2.5em}
\end{table}

Table~\ref{tab:binary} reveals that B4DL-Ego has a lower false-positive rate than B4DL on binary existence questions, but also lower positive recall. The reduction in false assertions outweighs the increase in missed positive claims, yielding accuracies of 67.20\% and 66.30\%, respectively. False assertions are more frequent on position questions, where false-positive rates are 78.40\% for B4DL and 74.80\% for B4DL-Ego. 

\subsection{RQ3: Do inference-time changes repair errors?}
\label{sec:results_mitigation}
\begin{table}[t]
\centering\footnotesize
\caption{Paired intervention results. $R/D$: repairs/regressions. Accuracy changes and 95\% scene-bootstrap intervals are in percentage points. Prompt denotes anti-position prompting.}
\label{tab:paired}
\begin{tabular}{@{}llrrrl@{}}
\toprule
Model & Intervention & $R$ & $D$ & $\Delta A$ & 95\% interval\\
\midrule
B4DL & Prompt & 58 & 65 & -0.07 & [-0.27, +0.13]\\
B4DL & $\alpha=0.5$ & 117 & 114 & +0.03 & [-0.25, +0.31]\\
B4DL & $\alpha=1$ & 119 & 115 & +0.04 & [-0.25, +0.34]\\
B4DL & $\alpha=1.5$ & 119 & 116 & +0.03 & [-0.27, +0.33]\\
B4DL-Ego & Prompt & 50 & 117 & -0.67 & [-0.93, -0.41]\\
B4DL-Ego & $\alpha=0.5$ & 161 & 146 & +0.15 & [-0.22, +0.52]\\
B4DL-Ego & $\alpha=1$ & 161 & 143 & +0.18 & [-0.19, +0.54]\\
B4DL-Ego & $\alpha=1.5$ & 161 & 142 & +0.19 & [-0.19, +0.55]\\
\bottomrule
\end{tabular}
\vspace{-2em}
\end{table}

Relative to default decoding, TSCD accuracy gains are 0.03--0.04 percentage points for B4DL and 0.15--0.19 for B4DL-Ego (Table~\ref{tab:paired}).  All six 95\% scene-bootstrap confidence intervals include zero. 
For B4DL-Ego at $\alpha=1.5$, 161 errors are corrected but 142 previously correct answers become incorrect, leaving a net gain of 19 out of 10,000 questions.
The dominant failures are unchanged: temporal answers remain A and lateral-motion answers remain No.

Anti-position prompting also fails to improve aggregate accuracy and reduces response validity for B4DL-Ego. An instruction to avoid positional preference does not supply the missing distinction between candidates. Since this instruction changes only multiple-choice prompts, variation in binary answers reflects the separately sampled runs rather than a change to those prompts.

In Figure~\ref{fig:qualitative}, TSCD changes an incorrect spatial answer to another incorrect answer in (B), preserves a correct non-A answer in (C), and leaves the incorrect temporal choice unchanged in (D). The motion pair in (E,F) illustrate the distinction between rejecting a false relation. 

\subsection{Scoring and indexing checks}
\label{sec:checks}
Strict parsing changes fewer than 0.1\% of responses and leaves the dominant answer patterns unchanged. 
Saved reference indices and temporal permutations are valid for the recorded feature lengths. The dominant response patterns therefore persist after correcting the identified scoring discrepancies.

\section{Discussion and Limitations}
\label{sec:discussion}
\noindent\textbf{What the diagnostics add.}
The observed errors do not follow a single preference for ``No'' or ``A''. Multiple-choice responses favor A, lateral-motion questions receive No, and position questions often elicit false Yes answers. This task dependence matters when comparing checkpoints: a higher average can result from rejecting more false claims while detecting fewer true relations. The protocol identifies these trade-offs and the relations on which they occur. For driving-related questions, separate measures of missed objects, false assertions, and paired discrimination indicate what needs to improve, rather than treating all accuracy gains as equivalent.

\noindent\textbf{Reference quality and evaluation scope.}
The benchmark evaluates two configurations from one model family and inherits its temporal intervals from B4DL. Sparse returns, near-tied distances, and boundary cases warrant an independent annotation and observability audit. The motion prompts use qualitative words with threshold-based labels, so small directional changes can receive negative references. A threshold-explicit prompt evaluation would help quantify this ambiguity. Broader model coverage, option permutations, and same-checkpoint tests with and without LiDAR would further separate answer-interface effects from failures to use geometric evidence. The current experiments concern question answering, rather than closed-loop driving performance.

\noindent\textbf{Temporal contrast and correction.}
TSCD rewards clean--shuffled disagreement, which helps only when that disagreement favors the correct answer. A logit component shared by both branches remains in Eq.~\eqref{eq:tscd}, so temporal contrast alone does not remove a shared answer preference. Testing this explanation requires branch logits and matched decoding runs. In particular, $\alpha=0$, identity-permutation controls, and repeated sampling seeds would isolate the contrastive term from differences in the decoding loop and sampling.

\section{Conclusion}
\bench{} combines explicit reference-answer rules with diagnostic evaluation of LiDAR language models. Across the recorded responses, fixed-option predictions obscure weak multiple-choice performance, identical questions often fail to distinguish opposite scene relations, and true lateral-motion relations are consistently missed. The tested inference-time interventions leave these failures largely unchanged. Fixed-answer controls, paired questions, relation-specific recall, and repair counts provide a more informative account of spatio-temporal reasoning than aggregate accuracy alone.

\noindent\textbf{Generative AI use disclosure.}
OpenAI Codex was used to assist with drafting and debugging portions of the experimental code and with language editing throughout the manuscript. The authors reviewed and validated the resulting code and text. All reported results were obtained from actual experiment runs and verified by the authors; no empirical result values were invented, altered, or synthesized by AI.

\balance
\bibliographystyle{IEEEtran}
\bibliography{references}
\end{document}